\documentclass{article}

    \usepackage[preprint, nonatbib]{neurips_2023}

\usepackage[utf8]{inputenc} 
\usepackage[T1]{fontenc}    
\usepackage{url}            
\usepackage{booktabs}       
\usepackage{amsfonts}       
\usepackage{nicefrac}       
\usepackage{microtype}      
\usepackage{xcolor}         

\usepackage{amssymb}
\usepackage{makecell}
\usepackage{hyperref}
\usepackage{amsmath}
\usepackage{algorithm}
\usepackage{algpseudocode}
\usepackage{algcompatible}
\usepackage{graphicx}
\usepackage{subfigure}

\title{SORTAD: Self-Supervised Optimized Random Transformations for Anomaly Detection in Tabular Data}

\author{%
  Guy Hay \\
  AI Solutions Group\\
  Intel Corporation\\
  \texttt{guy.hay@intel.com} \\
  \And
  Pablo Liberman \\
  AI Solutions Group\\
  Intel Corporation\\
  \texttt{pablo.liberman@intel.com}
}

\begin{document}

\maketitle

\begin{abstract}
  We consider a self-supervised approach to anomaly detection in tabular data. Random transformations are applied to the data, and then each transformation is identified based on its output. These predicted transformations are used to identify anomalies. In tabular data this approach faces many challenges that are related to the uncorrelated nature of the data. These challenges affect the transformations that should be used, as well as the use of their predictions. To this end, we propose SORTAD, a novel algorithm that is tailor-made to solve these challenges. SORTAD optimally chooses random transformations that help the classification process, and have a scoring function that is more sensitive to the changes in the transformations classification prediction encountered in tabular data. SORTAD achieved state-of-the-art results on multiple commonly used anomaly detection data sets, as well as in the overall results across all data sets tested.
\end{abstract}

\section{Introduction}

Anomaly detection \cite{chandola2009anomaly} is the process of detecting samples that do not align with the majority distribution of the studied population. In real applications of fault or fraud detection \cite{phua2010comprehensive, garcia2009anomaly}, these anomalies can point to defects or fraudulent samples. Therefore, there is an ever growing need for improving anomaly detection algorithms. Anomalies are often not well represented in the training data, even if they are labeled, due to a small number of samples from each possible anomaly source. In these cases it is not possible to train supervised methods to classify normal vs. abnormal samples. Unsupervised methods that only learn the normal samples distributions address these challenges and are preferred in anomaly detection.

Many anomaly detection methods have been proposed. Most methods rely on statistical properties like the density of the samples \cite{breunig2000lof}, or finding a general space that encapsulates the normal samples as in \cite{scholkopf2001estimating, ruff2018deep}. Another promising method is self-supervision. Self-supervision is an unsupervised learning paradigm, in which a supervised model solves one or more secondary tasks, based on constructed labels that are derived from the features themselves. The trained model is then used to tackle the primary task. An intuitive example is using Auto-Encoders for anomaly detection \cite{chalapathy2019deep, vincent2008extracting}. In this case the primary task is finding anomalies, while the secondary task is learning to compress the data, using the features themselves as the labels. The compression reconstruction error of each sample is used as the sample's anomaly score. The samples with the highest scores are then considered anomalies. Self-supervision doesn't rely on the sample labels, and usually works best when using only the normal samples for training, and was shown to improve model robustness \cite{hendrycks2019using}.

Recent advancements in self-supervised learning for anomaly detection, including GEOM \cite{golan2018deep}, GOAD \cite{bergman2020classification}, $SLA^2P$ \cite{10.1145/3511808.3557697}, $E^3Outlier$ \cite{wang2019effective} and RecTrans \cite{hu2020advanced}, all used one new approach with the same main assumption. They randomly transform the samples, and then predict for each transformed sample which transformation was applied to it. The assumption is that anomalous samples will display some irregular behavior following transformations, while "normal" samples will behave similarly following such transformations. Thus, a classifier that is trained to detect from transformed data samples their corresponding transformation will have better accuracy and higher confidence for normal samples, as opposed to abnormal samples.

This approach was first introduced in GEOM \cite{golan2018deep} and in \cite{komodakis2018unsupervised} for images. GEOM uses the images natural symmetry for rotations to randomly rotate images, and then predicts the applied rotation with a neural network. It then scores each sample using Dirichlet distribution approximation normality score.

Anomaly detection based on random transformations in tabular data faces some obstacles that don't exist in images. The first obstacle, which we name "Unrelated", is that tabular data has no underlying relation between the features. Therefore, there are no symmetries that can be considered independent of differences between samples, like rotations in images as in GEOM \cite{golan2018deep}; \cite{komodakis2018unsupervised}. These symmetries were used in the selection of transformations in order to avoid mixing the transformations outputs.

The second obstacle, "Non-Sparse", is that for tabular data the samples tend to be concentrated in small parts of the feature space (a relatively small L2 distance between them). This is clearly the case when normalization is used, making samples typically reside within the unit ball. This creates a need for transformations to be very different from one another, to enable a classifier to distinguish between the transformations applied. Otherwise transformations will map samples to the same space. This is unlike the case of images, where mapping different images to the exact same image by rotation is unlikely, even if they contain the same object, but with different backgrounds for example.

The third obstacle in tabular data, "Easy Detection", is that anomalies can be separated from normal samples by a certain way that amplifies the transformations detect-ability. This creates a problem for methods that rely on the difficulty of identifying the transformation as a measure for anomaly (e.g. RecTrans \cite{hu2020advanced}). An intuitive explanation is with the simple transformations of $t_1(x)=x^3$ and $t_2(x)=x+0.3$. For values larger than 1, $t_1$ will usually produce larger values than $t_2$. If most normal values are around 1.2, for example, then the classifier may learn to classify high values to $t_1$. This may cause a problem if abnormal values are much higher. Even though these values are abnormal, the classifier will be very successful in classifying them, since it learned to classify high values to $t_1$. These cases will cause the summation scoring method in RecTrans, which sums just the correct transformations predictions, to fail.

In the tabular data domain both GOAD \cite{bergman2020classification} and RecTrans  \cite{hu2020advanced} use random transformations. GOAD uses affine transformations, and then feeds the results to a neural network \cite{neural_network} trained with triplet loss \cite{he2018triplet}. This increases the outer variance, i.e., the mean distance between samples transformed by different transformations, while decreasing the inner variance, the variance of the resulting samples with the same transformation. It then proceeds to score each sample with the sum of the normalized Gaussian probabilities predicted for each transformation. RecTrans, on the other hand, uses random polynomial transformations. It then predicts the applied transformation with a neural network. It scores each sample with the sum of the correct predictions of the sample (the Summation Scoring method). The randomization of RecTrans introduces high variance in the results. GOAD is less susceptible to this variance since it optimizes the transformations using the neural network. 

To address the obstacles described above, we propose SORTAD, a novel algorithm that is tailor maid to solve these challenges. Unlike the previous methods, it selects from the random transformations generated the ideal transformations that would help the classification process. The selected transformations are those that better separate the transformed data from the data transformed by the previously selected transformations. In addition, SORTAD proposes a scoring function that is especially sensitive to the changes in the transformation classifiers predictions encountered in tabular data, as described below. We show that SORTAD outperforms previous methods on commonly used anomaly detection benchmarks.

We summarize the new contributions in SORTAD compared to previous work:
\begin{itemize}
  \item We present a method to select transformations from randomly generated transformations, to improve overall anomaly detection and reduce performance variance, while using less transformations (leading to less computational overhead). We thus address the Unrelated, Non-Sparse and Easy Detection obstacles described above. 
  \item We present a modified version of the scoring method used in GEOM, specifically designed for anomaly detection in tabular data. This addresses the Easy Detection challenge.
  \item We present a modified version of the reversible polynomial transformation, introduced in RecTrans, which is more numerically stable.
\end{itemize}

The remainder of this paper is structured as follows: Section 2 provides a detailed description of the proposed algorithm. Section 3 provides a description of the model evaluation method used which is specifically designed for anomaly detection in production environments, that doesn't rely on knowing the amount of anomalies in the dataset. We also highlight its benefits over F1 score. Section 4 specifies the experimental setting and presents and discusses the results. Lastly, Section 5 is an overall discussion.

\section{SORTAD Method}
The underlying assumption of our method, as in previous methods mentioned in the Introduction, is that anomalous samples will display irregular behavior following transformations, while normal samples will behave similarly following such transformations. 

Other than the empirical evidence in the results previous methods, we believe the main assumption has theoretical intuition as well. Under the assumptions that the features present are enough to detect anomalies, a reasonable assumption, there is an underlying reason that these samples are anomalies. Meaning in some underlying way they are different than the normal samples. Camparing the behavior of normal samples to newly seen samples should show which samples are anomalous.

Based on this assumption, SORTAD randomly transforms the data using reversible polynomial transformations, introduced in RecTrans \cite{hu2020advanced}. These reversible polynomial transformations are as follows:
\begin{equation}
y_1 = x_{p_1} + G(y_{p_2})
\label{eq_reversible_polynomial_transformations_1}
\end{equation}
\begin{equation}
y_2 = x_{p_2} + F(x_{p_1})
\label{eq_reversible_polynomial_transformations_2}
\end{equation}
Where $x\in \mathbb{R}^p$ is the input; $p_1$ and $p_2$ are randomly non intersecting feature lists of the same length upholding $|p|=|p_1|+|p_2|$; $x_{p_1}$ and $x_{p_2}$ are $p_1$ and $p_2$ corresponding sample feature values; $(G)_i: \mathbb{R}\xrightarrow{}\mathbb{R}$ and $(F)_i: \mathbb{R}\xrightarrow{}\mathbb{R}$ are polynomial transformations, which are applied element wise to each feature value $i$. The final transformed sample is a concatenation of $y_1,y_2$. In cases where there is an uneven amount of features, a zeros vector is added to the feature space. The reversibility of the transformations is proven in \cite{hu2020advanced}, thus, guarantees that with an optimal classifier it is possible to classify each transformation by reversing the input.

Similarly to RecTrans, $F$ and $G$ use different polynomial bases, Chebyshev and Legendre to increase their uniqueness. In addition, applied one after another, limited the amount by a hyper-parameter. These transformations help address the Non-Sparse and Unrelated obstacles.

Applying several transformations in a row introduces two crucial problems: underflow and overflow of the feature space. Overflow happens when a value gets too high for the current computer representation to handle, making the algorithm crash. Underflow happens when two values of very different magnitudes are summed, causing the smaller value to not be represented in the resulting number. This causes several samples to have the same prediction. Both cases are devastating for the detection performance. To address this problem two main elements were added to the transformations:
\begin{itemize}
  \item Divide factor – transformations were added a constant divider that is $10^{(d-h)}$, where $d$ is the degree of the polynomial, and $h$ is a hyper-parameter (usually 2). This change helps prevent overflow, especially in data sets normalized with robust scaling \cite{scikit-learn_robust}. Robust scaling is a scaling technique widely used for anomaly detection data sets, which scales the data using the median value and the IQR of the feature, thus making it less sensitive to outliers (anomalies) in the data than standard scaling. 
  \item Non-Constant – Chebyshev and Legendre polynomials have in every second base element a constant addition, when represented in the ordinary polynomial basis. The constant value causes underflow because it is added to the low values normally outputted from the rest of the polynomial equation. Thus, choosing basis elements that do not have a constant element decreases underflow.
\end{itemize}

To help solve the Unrelated, Non-Sparse and Easy Detection obstacles, we propose improving the transformation creation process by selecting the best transformation out of randomly generated transformations. Improving the transformation selection enables the use of less transformations. It improves the classifier's ability to distinguish between them and retain the transformations that best separate the anomalies behavior from the normal samples behavior. This boosts detection performance, in addition to reducing the computational resources needed. 

\begin{figure}[ht]
\begin{center}
\includegraphics[scale=0.45]{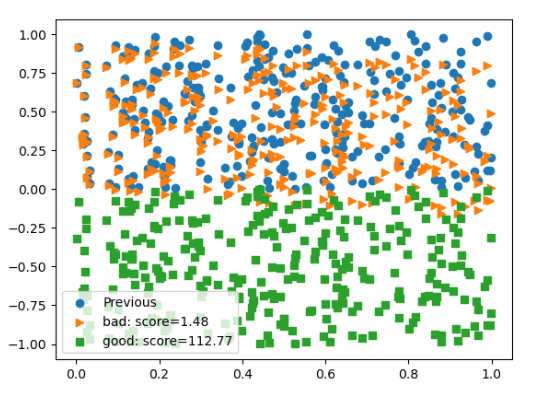}
\end{center}
\caption{An example of two transformations and their corresponding score using Eq. \ref{eq_modified_triplet_score} with the original data. The green square transformation result with a score of 112.77 can be easily separated from the original blue circle data. The orange triangle transformation result with a score of 1.48, is barely distinguishable from the original data. Causing the classifier to separate the green squares from the blue circles better than the orange triangles from them. Under the assumption that anomaly samples will display irregular behavior, anomalies under the green square transformations would be harder to separate than the normal data, unlike what will happen under the orange triangle transformations where all data normal and anomalous will be hard to separate.}
\label{fig_transformation_examples}
\end{figure}

To this end, in each step SORTAD randomly creates $k$ temporary transformations, assigns each transformation a score based on the resulting structure of the transformed samples, and then chooses the best transformation based on this score until $M$ transformations are chosen. The scoring function Eq. \ref{eq_modified_triplet_score} is proposed. The main idea is to help the transformation classification algorithm by forcing the transformed samples to be farther away than the results of the previous transformations outputs center of masses (higher outer variance). In addition, the resulting samples should display lower inner variance (mean distance from the center) allowing further separation from previous transformations outputs. Since both the center of mass and the inner variance are guided by normal samples, anomalous sample's, if present, affect are limited. More importantly, these metrics best captures the normal samples behavior and not the anomalous. Since we assume that anomalous samples will display irregular behavior under transformations they will probably be farther away from the center of mass and therefore close to previous transformations centers. Causing the classifier to give poor predictions to them, allowing the anomaly score to be high for anomaly samples. Moreover, since it strives to find transformations with low inner variance it facilitate our assumptions that normal samples will behave the same under transformations, since we choose the transformations that they do. Thus, choosing low inner variance helps strengthen our main assumption and facilitates anomaly detection. Our proposed score is a modification of the triplet loss objective \cite{he2018triplet}. The modified score is described in Eq. \ref{eq_modified_triplet_score}.
\begin{equation}
tscore_m = \beta \sum_i |x_i-c_m| -(1-\beta)\sum_i min_{m'\neq m} |x_i-c_{m'}|
\label{eq_modified_triplet_score}
\end{equation}
Where $c_m$ is the center of transformation $m$ output, $m'$ is an index running on all previous transformations, $\beta$ is a hyper-parameter that is a weighting factor for the inner and outer variance’s importance. We use L1 norm, which is more robust to outliers. Fig. \ref{fig_transformation_examples} shows a qualitative example of two transformation and their corresponding scores. It shows that generating bad transformation can hurt the transformation process by generating transformations who's outputs are too similar to former transformations. Thus, not allowing the classifier to be able to classify normal samples correctly. Avoiding this, is one of SORTAD's biggest advantages.

\begin{figure*}[ht]
\begin{center}
\includegraphics[width=\textwidth]{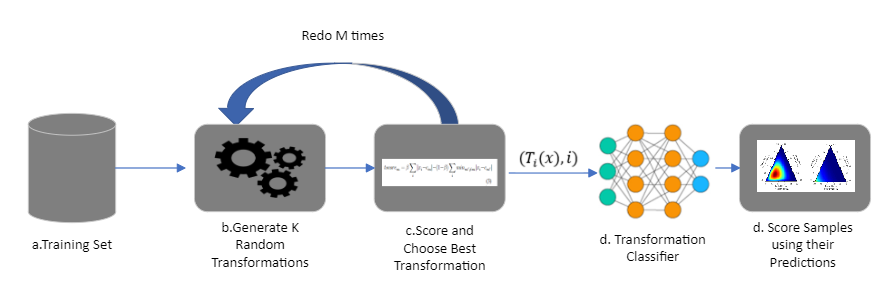}
\end{center}
\caption{An overview of SORTAD. \textbf{a.} Training set. \textbf{b.} Generates K temporary transformations. \textbf{c.} Scores the K temporary transformations using Eq. \ref{eq_modified_triplet_score}, and chooses the best transformation. b and c happens M times resulting in M transformations. \textbf{d.} Transformation classifier, classifies trains to classify the transformations using their corresponding outputs. \textbf{d.} The scoring function. Uses Eq. \ref{eq_modified_normality_score} to score the samples using the transformations predictions.}
\label{fig_framework}
\end{figure*}

After $M$ transformations are selected the outputs of the transformations and their corresponding labels are inputs for the transformation classifier. The transformation classifier receives as input a transformed sample and predicts the applied transformation. An overview of the model is given in Fig. \ref{fig_framework}.

After the training process the prediction of the classifier is used for scoring. The Summation Score, in which each sample's score is the sum of the correct predictions of the sample, was found to be prone to Easy Detection and gives anomalies high scores. This is the main reason for the fast drop in performance RecTrans demonstrated in Fig. \ref{fig_different_scoring_methods}. The Dirichlet probability scoring method proposed in GEOM, described in Eq.\ref{eq_dirichlet_probability_function}, helps address this problem. It considers when a probability score of the transformation is different from what was seen for the training samples, and uses the predictions for other transformations. This normality score computes the probability of a sample's transformation prediction scores in the Dirichlet distribution space approximated from the classifier's prediction on the training samples. The final score for each sample is the sum of probabilities of the classifier's prediction for each transformation on the approximated Dirichlet distribution. This also helps solve the problem where predictions of the applied transformation for anomalous samples and normal samples receive the same score, but the scores on the wrong transformations are significantly different. Thus, for the Summation Score, both anomalies and normal samples will receive the same score, but in the Dirichlet probability score, anomalies will receive a lower score. Additionally, by considering all the transformations predictions for given transformations, more information is saved and used to determine which sample is anomalous.

\begin{equation}
n(x) = \sum_{m=0}n_m(x) = \sum_{m=0}\sum_{j=0}(\alpha_{m_j} - 1)log(y(T_m(x))_j)
\label{eq_dirichlet_probability_function}
\end{equation}

Where $\alpha$ is the concentration parameter for the Dirichlet probability approximated from the training samples, $T$ is transformation function, $m$ is an index for all transformations, $j$ is an index for all transformation prediction scores. This means that $m_j$ is the index for transformation $j$ when transformation $m$ is applied. Normal samples scored a higher value than anomalies, as described below.

\begin{figure}[ht]
\begin{center}
\includegraphics[scale=0.45]{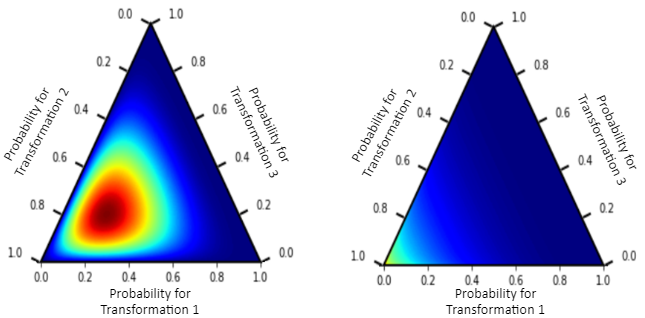}
\end{center}
\caption{On the left, Dirichlet distribution PDF values, without experiencing Easy Detection with $\alpha=[4,2,2]$. On the right, while experiencing Easy Detection with $\alpha=[4,0.95,0.95]$. High values are in red, and low are in blue. Note that the distribution is skewed to the vertex, causing all values near that vertex to have very high probability, in turn causing anomalies that have abnormally high probabilities to have higher values. Whereas in the normal case (on the left), this does not happen, and values that are highly certain (closer to the vertex), have lower probabilities.}
\label{fig_dirichlet_distribution_probabilities}
\end{figure}

For the Dirichlet probability score, Easy Detection persists. This is in contrast to the intuition that the scores for anomalies are abnormally high, therefore they should get low probabilities. The reason for this occurrence lies in the Dirichlet probability score, which is not sensitive enough for the change. The $\alpha$ parameter of the Dirichlet distribution, in Easy Detection transformations will have a very high value for the transformation and very low values for the rest. This causes the distribution to be centered at the edge, where the correct transformation gets a prediction score close to 1. Fig. \ref{fig_dirichlet_distribution_probabilities} shows the Dirichlet distribution for different $\alpha$ values showing how the distribution gets skewed to the vertex. As a result, every sample that gets a near perfect prediction score, gets a very high normality score regardless of how high. In addition, very low prediction values have a very large log value in absolute terms, causing the normality scores to be governed by the predictions for the wrong transformations, because they occur with $\alpha_{m_j} < 1$. Since in these cases anomalies get a higher correct prediction score, their wrong prediction score is lower, therefore, their normality score is higher. They are so high that the modification needed is to use the distance from the mean score on the training samples. The final modified normality score is in Eq. \ref{eq_modified_normality_score}.
\begin{equation}
n_m(x) = 
\begin{cases}
n(x), & \min\limits_{j}(\alpha_j) \geq 1\\
|n(x)-\overline{n}(x)|, & \min\limits_{j}(\alpha_j) < 1 \land n(x) \geq 0 \\
|n(x)-\overline{n}(x)| * R, & \min\limits_{j}(\alpha_j) < 1 \land n(x) < 0
\end{cases}
\label{eq_modified_normality_score}
\end{equation}
Where $\overline{n}(x)$ is the mean of Eq. \ref{eq_dirichlet_probability_function} on the training data. The last option in Eq. \ref{eq_modified_normality_score} is to account for the information loss while using the distance from the mean score. Since with Easy Detection all samples are expected to have positive results, samples that still have negative results are more likely to be anomalies. Therefore, a multiplication factor is added to incorporate this information into the score. For all the test results, R=3 was used. The higher the score the more "normal" the sample is.

\section{Model Evaluation Method}
Current metrics for anomaly detection use the well known F1-score with the anomaly percentage in the training data. Thus, they rely on having the similar anomalies percentage in both training, validation and test sets. In addition, for this evaluation metric to be accurate for production purposes, it is necessary for the algorithm to have a high enough number of samples. Otherwise, it is unclear where the split between normal and anomalous occurs. That is, if the anomaly percentage is 1.45\% and only 100 samples are in the predicted set, it is unclear whether 2 or 1 samples should be considered anomalies. These assumptions don’t apply in many real anomaly detection applications, such as fault or fraud detection, which use a pre-computed threshold in the anomalies scorer to label samples as normal or abnormal. This threshold is computed offline on the train or validation set to achieve the desired balance between false negative and false positive errors.

To evaluate the models performance in these systems, we propose to use the the recall ratio vs a random classifier in addition to ROCAUC, and Recall metrics at predefined thresholds. The “VS Random” metric, is the ratio between the recall in the actual fraction of samples labeled as anomalous in the evaluated set using a threshold value determined by the training set and the recall a random classifier would have achieved in that same actual fraction. Unstable thresholds between train, validation and test set, pose a problem in production. One algorithm's recall can be significantly higher than the other only because it’s learned thresholds are very different than those that should have been used, if learned on the different set. For example, if the 5\% threshold score is used, it may classify 20\% of the newly tested samples as anomalies. In this case, the "actual alert percentage" is 20\%. This usually results in a higher recall but a higher percentage alerted, which is not favorable. The recall score without adjusting to the actual alert percentage will be called the "Non Adjusted Recall". In these cases the VS Random score will be much lower. In addition, algorithms that have a close actual alert percentage to the desired percentage, show that the underlying distribution of the training data was learned, which gives a higher confidence of its stability and generalization.
We emphasize that the VS Random metric is the most important metric for anomaly detection, since it best describes the amount of target samples eliminated. For example a VS Random score of 2 means that the algorithm eliminated 2 times more anomalies than a random classifier would. For imbalance data sets it is common to use two thresholds the real desired threshold and a larger threshold to evaluate the stability of the model. In our evaluation 3\% and 10\% were used as the main and stability percentages respectively.

\section{Experiments}
\textbf{Datasets:} All data sets where randomly split to train and test sets, with data sets containing more than 30 anomalies and enough samples split into train, validation and test sets. The split was done while preserving the anomaly rate between sets. When all three sets exist, the algorithms are trained on the training set, then hyper-parameters are tuned on the validation set and the training set and then evaluated on the test set. Hyper parameters tuned on the training and validation set show higher stability and are less likely to overfit. In cases were no validation set exists, the hyper-parameter tuning was done only on the training set. Evaluation was done on 10 predetermined data sets: Mammography, SMTP, Forest Cover, Thyroid, Shuttle, Pendigits, Arrihythmia, Annthyroid, Vowels \cite{Rayana:2016}, and Credit card \cite{credit-card}. The data sets are described in the Appendix. These data sets are commonly used to evaluate many anomaly detection algorithm and are from various domains and have a diverse number of samples and features. We focus on tabular data with lower dimension as opposed to image data sets since, we believe they better represent the challenges in anomaly detection in tabular data. All the data sets were scaled with robust scaling.

\textbf{Baseline Methods:} Classical baseline methods consist of: Isolation Forest \cite{liu2008isolation}, One-Class SVM \cite{scholkopf2001estimating}, Elliptic Envelope \cite{rousseeuw1999fast} and LOF \cite{breunig2000lof}. More recently developed algorithms, including state-of-the-art methods are: GOAD \cite{bergman2020classification}, Anomaly Detection for Tabular Data with Internal Contrastive Learning, (ICL) \cite{shenkar2022anomaly}, COPOD \cite{li2020copod}, and RecTrans \cite{hu2020advanced}. For the classical baseline methods the default hyper-parameters in the python package scikit-learn \cite{scikit-learn} were used. For ICL the non hyper-parameterized function was used with 5 different random seeds. For GOAD the hyperparameter search included 18 different hyper-parameter sets, consisting of a grid search of all hyper-parameters used in the paper with 1 epoch, including 3 random seeds, amounting to 54 different executions in total. For RecTrans 16 hyper-parameter sets were tested, 8 of which were later used for SORTAD, and 4 were from the paper.Additionally, 3 random seeds were used, a total of 36 execution sets. The full hyper-parameters for each method are summarized in the Appendix.

\subsection{Overall Scores}

To show SORTAD outperforms the baseline methods mentioned at the beginning of the section, the mean and standard deviation of VS Random at 3\%,10\% and the Non Adjusted Recall at 3\%,10\% of the results on the 10 data sets are presented in Table \ref{table_dataset_result_summary}. SORTAD outperforms all baseline methods on all metrics except for Non Adjusted Recall, where ICL has a higher result, but has a low VS Random which is significantly lower. This shows that its actual alert rate is a lot higher than the desired alert rate. Meaning that it just alerts on more samples, thus obtaining a high recall, indicating that SORTAD outperforms it. In addition, ICL showcased slower run times than SORTAD by an order of magnitude. SORTAD in general outperforms the second best model by 23\% and 28\% the third best. RecTrans was run with MinMax scaling instead of robust scaling because it crashed with almost all hyper-parameters used. This illustrates the importance of the modification SORTAD has, Divide Factor and Non-Constant degree, to the polynomial transformation. Scores for each individual data set are in the Appendix. In addition, summary of individual results and ROCAUC scores can be found in the Appendix. SORTAD outperforms all models in the VS Random 10\% as well but in a smaller percentage. This is expected as 10\% as it is usually higher than the anomaly rate, meaning that it is too high to solely depict a models performance as the differences are averaged out due to the high threshold cutoff. Moreover, it is used to prove the stability of the model but rarely used as the actual sampling rate. As evident, that only Arrihythmia has more than 10\% positive negative rate.

\begin{table*}[ht]
\centering
\caption{The mean and standard deviation of VS Random at 3\%,10\% , and the Non Adjusted Recall at 3\%,10\% of all data sets tested. SORTAD outperforms on all metrics except for Non Adjusted Recalls, where ICL has better results and SORTAD is second, but has VS Random 3\% and 10\% is significantly lower, it shows that its actual alert rate is a lot higher than desired, indicating that SORTAD outperforms it. In addition, ICL is slower by a order of magnitude. SORTAD on average outperforms the second best model by 23\% and 28\% the third best.}
\resizebox{\textwidth}{!}{
\begin{tabular}{|ccccc|}
 \toprule
 Model & VS Random 3\% & VS Random 10\% & Non Adjusted Recall 3\% & Non Adjusted Recall 10\% \\ 
 \hline
 \midrule
 Isolation Forest & $13.50\pm6.26$ & $6.70\pm2.03$ & $0.402\pm0.232$ & $0.679\pm0.241$  \\ \hline
 One Class SVM & $13.85\pm6.77$ & $7.43\pm2.33$ & $0.413\pm0.207$ & $0.738\pm0.223$  \\ \hline
 Elliptic Envelope & $10.54\pm7.79$ & $5.38\pm3.00$ & $0.332\pm0.235$ & $0.577\pm0.293$  \\ \hline
 LOF & $13.80\pm8.49$ & $6.05\pm2.73$ & $0.513\pm0.264$ & $0.739\pm0.220$  \\ \hline
 GOAD & $12.90\pm7.16$ & $6.30\pm1.97$ & $0.401\pm0.235$ & $0.587\pm0.262$  \\ \hline
\makecell{ICL} & $12.15\pm8.37$ & $4.40\pm2.80$ & \textbf{0.632$\pm$0.271} & \textbf{0.833$\pm$0.218}  \\ \hline
 COPOD & $11.31\pm8.71$ & $6.01\pm2.48$ & $0.327\pm0.256$ & $0.605\pm0.256$  \\ \hline
 RecTrans (MinMax) & $14.43\pm10.96$ & $7.16\pm2.73$ & $0.389\pm0.234$ & $0.616\pm0.228$  \\ \hline \hline
 \textbf{SORTAD} & \textbf{17.74$\pm$7.64} & \textbf{7.65$\pm$2.49} & 0.526$\pm$0.256 & 0.748$\pm$0.246  \\ 
 \bottomrule
\end{tabular}
}
\label{table_dataset_result_summary}
\end{table*}

\subsection{Scoring Method Analysis}
First we demonstrate the problem of Easy Detection on SMTP dataset, which caused the need for a modified scoring function Eq. \ref{eq_modified_normality_score}. In Table \ref{table_sum_score_function_easy_detection} there are probability results outputted from a trained neural network for two normal and two anomalous samples. For all samples the neural network produced higher probabilities for the correct transformation but it is clear that for the anomalous samples, the probabilities are abnormally high. These high values will cause the sum scoring function to give poor results. Therefore, a scoring function as Eq. \ref{eq_modified_normality_score} that takes into account the outputted probability distribution of the predictions is needed. Since SORTADs modified scoring function will approximate the normal samples predictions distribution the two anomalies will receive a low probability that will in turn cause them to have a low normality score.

\begin{table*}[ht]
\caption{The probabilities given by the transformation classification network for normal and anomalous samples that were transformed by transformation 5. The high probabilities for transformation 5 show how Easy Detection will cause the summing scoring function to fail. Since SORTADs modified scoring function will approximate the normal samples predictions distribution which for this example will be near [0.18,0.17,0.21,0.19,0.25] the two anomalies will receive a low probability.}
\centering
\resizebox{\textwidth}{!}{
\begin{tabular}{|c c c c c c |} 
 \toprule
 Sample & T. 1 Probability & T. 2 Probability & T. 3 Probability & T. 4 Probability & T. 5 Probability\\ [0.5ex] 
 \midrule
 \hline
 Normal 1 & 0.187 & 0.164 & 0.227 & 0.177 & 0.246 \\ 
 \hline
 Normal 2 & 0.173 & 0.177 & 0.196 & 0.201 & 0.252 \\
 \hline
 Anomaly 1 & 1.11e-05 & 2.30e-05 & 3.55e-04 & 9.16 e-04 & 0.999 \\
 \hline
 Anomaly 2 & 1.92e-04 & 3.44e-04 & 2.56e-03 & 5.34e-03 & 0.992 \\
 \bottomrule
\end{tabular}
}
\label{table_sum_score_function_easy_detection}
\end{table*}

\begin{figure}[ht]
\begin{center}
\includegraphics[scale=0.37]{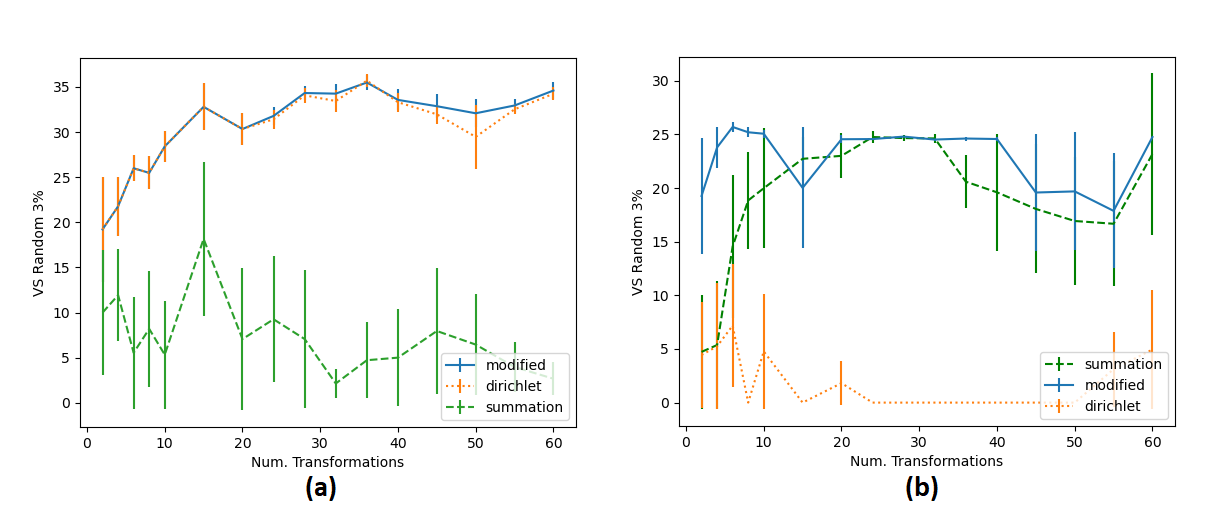}
\end{center}
\caption{Mean and standard deviation of VS Random at 3\% results for different scoring methods, as a number of transformations on \textbf{a.} Thyroid data \textbf{b.} SMTP data. While all other hyper-parameters were kept constant, including number of temporary transformations. The error lines are half the standard deviation, for visualizing purposes. Our modified scoring function (blue) improves the final score and stability in both cases. The two data sets each show when the Dirichlet (orange) or summation scoring function (green) fail, while such drop in performance doesn't occur for our method.}
\label{fig_different_scoring_methods}
\end{figure}

As mentioned in section 2, although the Dirichlet scoring function takes the outputted probabilities distributions into account it still fails in many cases. Fig. \ref{fig_different_scoring_methods} compares the VS Random at 3\% results of the Summation, Dirichlet and SORTAD scoring functions in Eq. \ref{eq_modified_normality_score}, as a function of the number of transformations. SORTAD scoring method outperforms the other methods, both in terms of mean result and standard deviation. Especially when more transformations are being used, as Easy Detection becomes more prominent. More importantly SORTAD's is less affected by the different data sets were the other scoring methods performance drops to zero, SORTAD's scoring method consistently performs well.

\subsection{Stability Analysis}

\begin{figure}[ht]
\begin{center}
\includegraphics[scale=0.35]{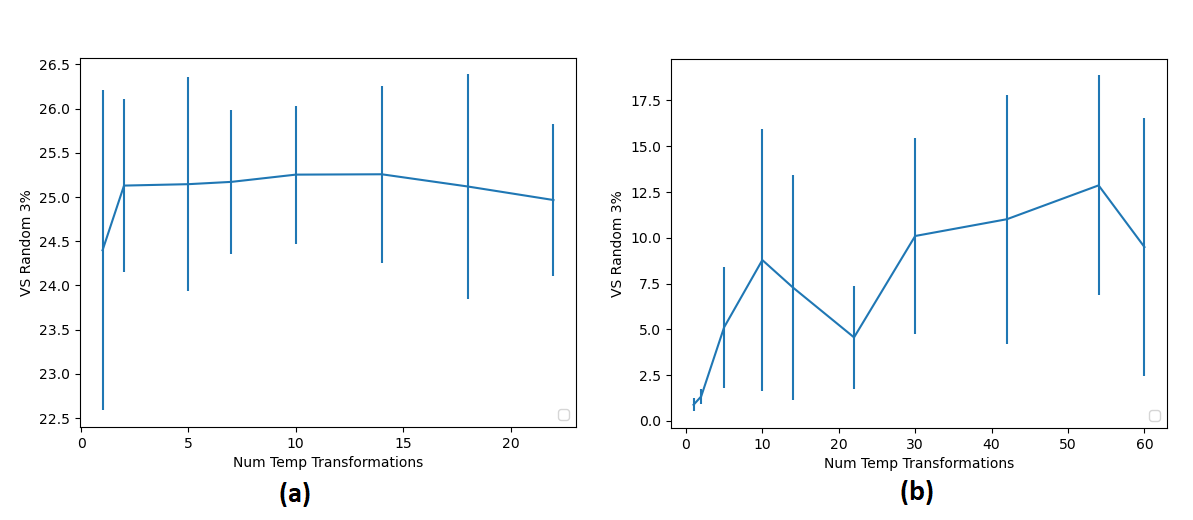}
\end{center}
\caption{Mean and standard deviation of VS Random at 3\% results as a function of the number of temporary transformations on the a. SMTP dataset b. Mammography data set. The error lines are the standard deviation. There is a clear performance gain in allowing SORTAD to choose from an increasing number of temporary transformations.}
\label{fig_different_num_temp_transformations_vs_random}
\end{figure}

To showcase the contribution of choosing the transformations, in Fig. \ref{fig_different_num_temp_transformations_vs_random} the VS Random at 3\% results of SORTAD with a different amount of temporary transformations is depicted. Increasing the number of temporary transformations allows the algorithm to have a bigger variety to choose from. Thus, allowing to evaluate Eq. \ref{eq_modified_triplet_score} ability to choose the correct transformations. Fig. \ref{fig_different_num_temp_transformations_vs_random}, shows the mean and standard deviation of the VS Random at 3\% results of 5 iterations of SORTAD with different seeds. First, there is a clear performance gain for choosing the transformations compared to randomly generating them. Showing that an ever growing number of temp transformations will result in better results. Thus, allowing SORTAD to choose from more temporary transformations is beneficial and makes the hyper parameter tuning significantly easier.

\section{Conclusions}

We propose a novel framework for anomaly detection named SORTAD: Self-supervised Optimized Random Transformation for Anomaly Detection. SORTAD's main innovation is managing to select the best transformations out of randomly generated transformations, thus increasing detection performance and stability while decreasing computational time. In addition, SORTAD utilizes a modified scoring method, specifically designed for tabular data, which was shown to produce better and more stable results while facing the Easy Detection problems encountered in tabular data. To conclude, SORTAD achieved state-of-the-art results on multiple anomaly detection data sets and in overall results on 10 anomaly detection data sets. SORTAD was tested using a validation set for hyper-parameters tuning, in cases where this is not possible from Fig. \ref{fig_different_num_temp_transformations_vs_random} and \ref{fig_different_scoring_methods} we can conclude that allowing SORTAD to use more transformations and to choose from more temporary transformation will usually yield better results.

\section{Acknowledgments}
We would like to express our sincere gratitude to Dr. Amitai Armon for his invaluable contributions to this research. His insightful comments and thorough review significantly enhanced the quality and clarity of this paper. Dr. Armon's expertise and thoughtful suggestions were instrumental in shaping the final manuscript. We are truly thankful for his time, dedication, and constructive feedback.

\bibliographystyle{unsrt}
\bibliography{main}

\section{Appendix}

\subsection{Datasets Description}\label{appendix_datasets}
\textbf{SMTP:} \cite{Rayana:2016} The detection of hazardous email, using 3 features. The feature names are not disclosed. The data set contains 95156 samples, out of which 30 anomaly samples amounting to 0.03\% of the samples. Splitted randomly to 2 sets, each containing 15 anomaly samples.\\
\textbf{Mammography:} \cite{Rayana:2016} The detection of mammography, using 6 features. The feature names are not disclosed. The data set contains 11183 samples, out of which 260 anomaly samples amounting to 2.32\% of the samples. Split randomly to, training: 4944 samples with 115 anomalies amounting to 2.33\%, validation: 2436 samples with 57 anomalies amounting to 2.33\%, and test: 3803 samples with 88 anomalies amounting to 2.31\%.\\
\textbf{Forest Cover:} \cite{Rayana:2016} The detection of anomaly type of forests, using 10 features. The feature names are not disclosed. The data set contains 286048 samples, out of which 2747 anomaly samples amounting to 0.96\%. Split randomly to, training: 95349 samples with 915 anomalies amounting to 0.96\%, validation: 95349 samples with 916 anomalies amounting to 0.96\%, and test: 95350 samples with 916 anomalies amounting to 0.96\%.\\
\textbf{Credit card:} \cite{credit-card} The detection of fraud in credit card transactions. The feature names are not disclosed, except for the "amount" feature which did not receive special attention. The data set contains 283726 samples, out of which 492 anomaly samples amounting to 0.17\%. Data has time ordering, therefore the split was done while keeping causality. Split to training: 132474 samples with 262 amounting to 0.20\% anomaly samples, validation: 60648 samples with 119 anomaly samples amounting to 0.20\%, test: 91685 samples with 111 anomaly samples amounting to 0.12\%. \\
\textbf{Thyroid:} \cite{Rayana:2016} The detection of thyroid, using 6 features. The feature names are not disclosed. The data set contains 3772 samples, out of which 93 anomaly samples amounting to 2.47\%. Split to training: 1257 samples with 31 anomalies amounting to 2.47\%, validation: 1257 samples with 31 anomaly samples amounting to 2.47\%, test: 1258 samples with 31 anomaly samples amounting to 2.46\%. \\
\textbf{Shuttle:} \cite{Rayana:2016} Using 9 features. The feature names are not disclosed. The data set contains 49097 samples, out of which 3511 anomaly samples amounting to 7.15\%. Split to training: 16365 samples with 1170 anomalies amounting to 7.15\%, validation: 16366 samples with 1171 anomaly samples amounting to 7.16\%, test: 16366 samples with 1170 anomaly samples amounting to 7.15\%. \\
\textbf{Pendigits:} \cite{Rayana:2016} Detecting from image attributes, hand written "0" from images of digits 1-9, using 16 features. The feature names are not disclosed. The data set contains 6870 samples, out of which 156 anomaly samples amounting to 2.27\%. Split to training: 2289 samples with 52 anomalies amounting to 2.27\%, validation: 2290 samples with 52 anomaly samples amounting to 2.27\%, test: 2291 samples with 52 anomaly samples amounting to 2.27\%. \\
\textbf{Arrihythmia:} \cite{Rayana:2016} The detection of cardiac arrhythmia, using 274 features. The feature names are not disclosed. The data set contains 452 samples, out of which 66 anomaly samples amounting to 14.60\%. Split to training: 226 samples with 33 anomalies amounting to 14.60\%, test: 226 samples with 33 anomaly samples amounting to 14.60\%. \\
\textbf{Annthyroid:} \cite{Rayana:2016} The detection of annthyroid, using 6 features. The feature names are not disclosed. The data set contains 7200 samples, out of which 534 anomalies amounting to 7.42\%. Split to three evenly distributed sets of 2400 samples each with 178 anomalies amounting to 7.24\%.\\
\textbf{Vowels:} \cite{Rayana:2016} Classifying outlier speaker using 12 features of the spoken time series. The data set contains 1406 samples out of which 50 are anomalies amounting to 3.43\%. Split to training: 485 samples with 17 anomalies amounting to 3.51\%, , validation: 485 samples with 17 anomaly samples amounting to 3.51\%, test: 469 samples with 17 anomaly samples amounting to 3.50\%.

\begin{table*}[ht]
\centering
\caption{Data set Summary, all data sets were split evenly to train/validation/test while preserving the amount of anomalies.}
\resizebox{\textwidth}{!}{
\begin{tabular}{|ccccc|}
 \toprule
 Data set & Number of Samples &  Number of Features & Number of Anomalies & Anomaly Ratio \\ 
 \hline
 \midrule
SMTP & 95156 & 3 & 30 & 0.03\% \\ \hline
Mammography & 11183 & 6 & 260 & 2.32\% \\ \hline
Forest Cover & 286048 & 10 & 2747 & 0.96\% \\ \hline
Credit Card & 283726 & 29 & 492 & 0.17\% \\ \hline
Thyroid & 3772 & 6 & 93 & 2.47\% \\ \hline
Shuttle & 49097 & 9 & 3511 & 7.15\% \\ \hline
Pendigits & 6870 & 16 & 156 & 2.27\% \\ \hline
Arrihythmia & 452 & 274 & 66 & 14.60\% \\ \hline
Annthyroid & 7200 & 6 & 534 & 7.42\% \\ \hline
Vowels & 1456 & 12 & 50 & 3.43\% \\
 \bottomrule
\end{tabular}
}
\label{table_smtp_result_summary}
\end{table*}

\subsection{Hyperparameters}\label{appendix_hyper_parameters}
All models used the following seeds: 1235, 7234, 3553. \\
\textbf{GOAD:} \cite{bergman2020classification} A mix of every combination of $Number of Rots \in \{64,256\}$, $Dimension Out \in \{32,64,128\}$, $NDF \in \{8,32,128\}$, all with 1 epoch, resulting in 18 different hyper-parameters. These are a mix of every combination of the hyper-parameters used in the paper except for the 25 epochs. \\
\textbf{RecTrans:}\cite{hu2020advanced} A mix of every combination of $Number of Transformation \in \{5, 10, 15, 22\}$, $Number of epochs \in \{1, 5, 20, 50\}$, $Transformation in a Row = 2$, $MaxPolynomialDegree=5$, Neural Network with 2 hidden layer the first with 64 nodes the second with 16, resulting in 16 different hyper-parameters. All the parameters used in the original paper and additional 12 more, 16 in total. \\
\textbf{SORTAD:} A mix of every combination of $Number of Transformation \in \{5, 10, 15, 22\}$, $Number of epochs \in \{1, 50\}$, $MaxPolynomialDegree=10$, $DivideFactor=2$, $NumberofTempTransformation=20$, Neural Network with 2 hidden layer the first with 64 nodes the second with 16, resulting in 8 different hyper-parameters.

\clearpage
\subsection{Individual Data Set Scores}\label{appendix_individual_scores}

\begin{table*}[ht]
\centering
\caption{VS Random 3\% results on specific data sets. The highest average rank is for SORTAD and is 2.8$\pm$2.1, second highest rank is for RecTrans and is 4.4$\pm$2.7. SORTAD on average outperforms the second best model by 23\% and 28\% the third best.}
\resizebox{\textwidth}{!}{
\begin{tabular}{|ccc ccc cccc|}
 \toprule
 Data set & \makecell{Isolation\\ Forest} & \makecell{One Class\\ SVM} & \makecell{Elliptic \\Envelope} & LOF & GOAD & \makecell{ICL} &  COPOD & RecTrans & \makecell{SORTAD\\ (ours)}\\ 
 \hline
 \midrule
 SMTP & 8.94 & 25.39 & 25.43 & 20.34 & 24.05 & 23.08 & 24.31 & \textbf{26.88} & 25.51  \\ \hline
 Mammography & 10.63 & 4.26 & 17.29 & 13.00 & 5.36 & 4.56 & 15.68 & 15.23 & \textbf{18.32}  \\ \hline
 Credit Card & 21.87 & 18.39 & 8.01 & 1.24 & 19.49 & 7.78 & 27.73 & \textbf{38.81} & 24.77  \\ \hline
 Thyroid & 17.04 & 23.67 & 20.29 & 18.26 & 18.81 & 5.49 & 4.27 & 9.19 & \textbf{25.93}  \\ \hline
 Forest Cover & 7.23 & 9.85 & 1.35 & 25.43 & 19.43 & 25.05 & 7.89 & 8.65 & \textbf{26.84}  \\ \hline
Arrhythmia & 5.14 & 5.87 & 5.02 & 1.83 & \textbf{6.85} & 3.42 & 4.57 & 4.57 & \textbf{6.85}  \\ \hline
Shuttle & 13.87 & 11.39 & 12.99 & 12.96 & 11.06 & \textbf{13.99} & 13.86 & 13.96 & 11.92  \\ \hline
Pendigits & 18.06 & 17.62 & 3.60 & \textbf{26.68} & 0 & 5.24 & 11.20 & 19.92 & 19.77 \\ \hline
Annthyroid & 7.35 & \textbf{11.09} & 9.05 & 6.82 & 10.79 & 8.36 & 3.60 & 7.07 & 10.37  \\ \hline
Vowels & 9.53 & 11.00 & 2.38 & 11.44 & 13.19 & \textbf{24.50} & 0 & 0 & 7.15
\\ \hline \hline
 mean$\pm$std & $13.50\pm6.26$ & $13.85\pm6.77$ & $10.54\pm7.79$ & $13.80\pm8.49$ & $12.90\pm7.16$ & $12.15\pm8.37$ & $11.31\pm8.71$ & $14.43\pm10.96$ & \textbf{17.74$\pm$7.64} \\ 
 \bottomrule
\end{tabular}
}
\label{table_vs_random_3}
\end{table*}

\begin{figure}[ht]
\includegraphics[width=\textwidth]{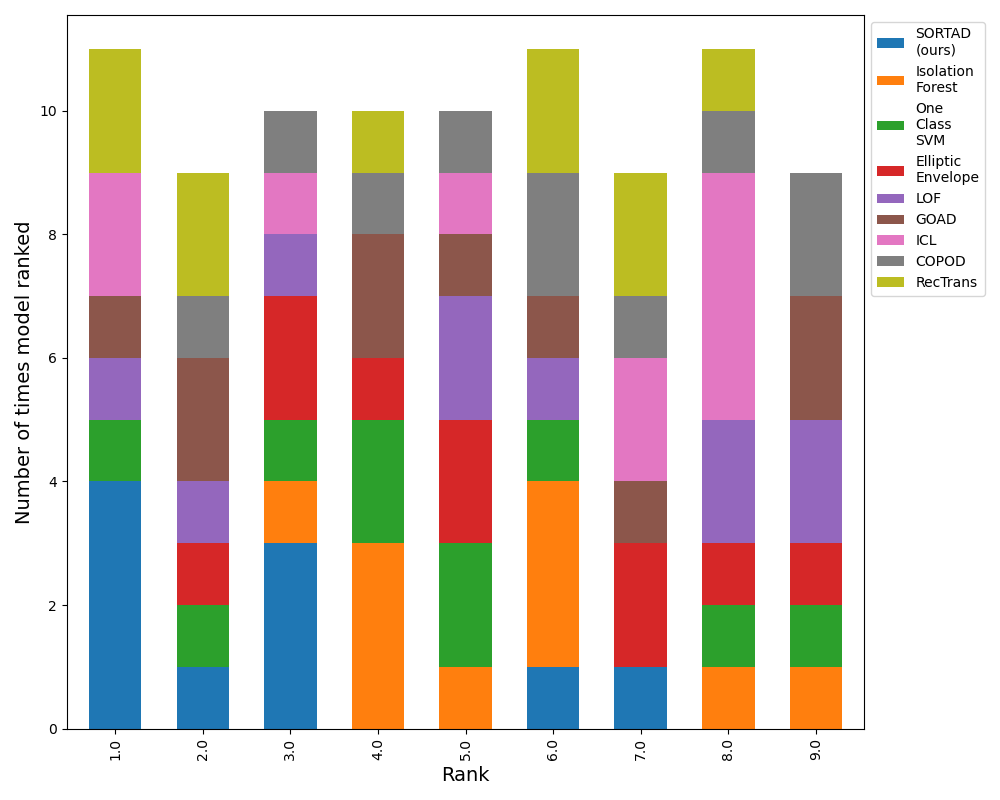}
\caption{Rankings of each model in the VS Random 3\% parameter. SORTAD in blue, more consistently occupies the leading spots. With an average rank of 2.8$\pm$2.1 SORTAD has the highest average rank, second highest rank is for RecTrans with 4.4$\pm$2.7. Note that some bars are above 10 indicating ties in the rankings. As 9 models are evaluated it is common that some models have uncharacteristically high result in a certain data set, thus adding noise to the rankings, evident by 3 models holding the first rank only in 1 data set, and poor results in the rest.}
\label{fig_ranking_with_rectrans}
\end{figure}

\clearpage
\begin{figure}[ht]
\includegraphics[width=\textwidth]{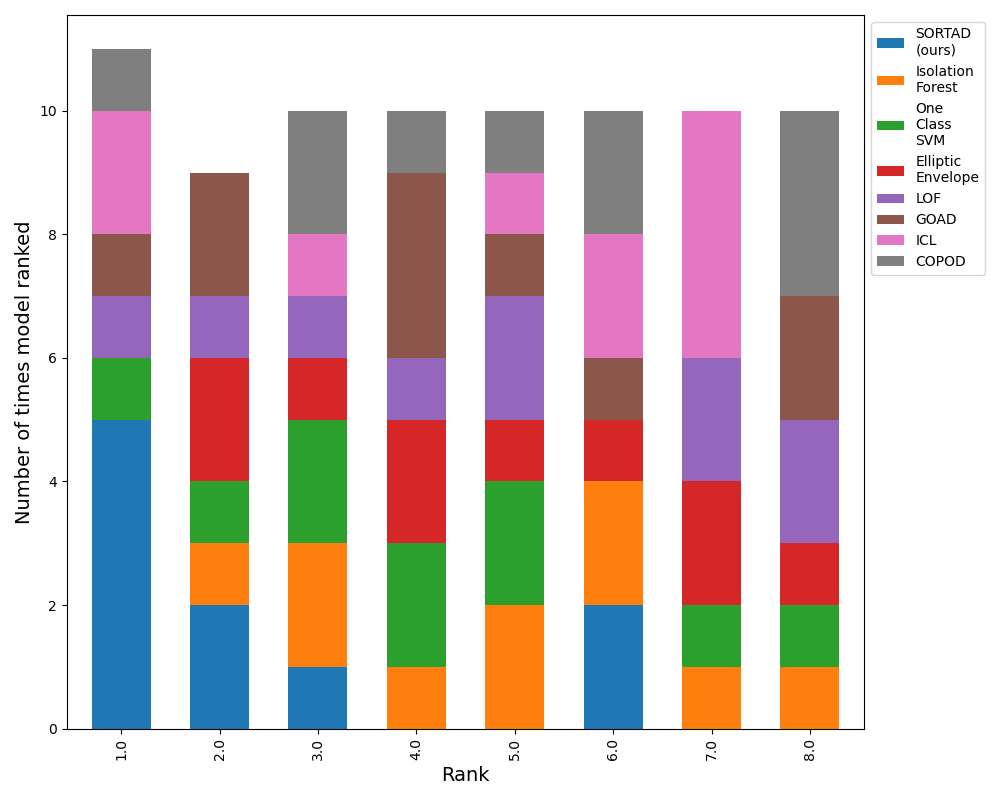}
\caption{Rankings of each model in the VS Random 3\% parameter. Without RecTrans which was normalized differently. This ranking best describes the "apples to apples" comparison between the models. SORTAD in blue, more consistently occupies the leading spots. Note that some bars are above 10 indicating ties in the rankings. As 9 models are evaluated it is common that some models have uncharacteristically high result in a certain data set, thus adding noise to the rankings, evident by 4 models holding the first rank only in 1 data set, and poor results in the rest.}
\label{fig_ranking_without_rectrans}
\end{figure}

\begin{table*}[ht]
\centering
\caption{VS Random 10\% results on specific data sets. The highest average rank is for SORTAD and is 3.0$\pm$1.7, second highest rank is for One Class SVM with 3.2$\pm$2.04. SORTAD outperforms all models in the VS Random 10\% as well but in a smaller percentage. This is expected as 10\% is usually higher than the anomaly rate, meaning that it is too high to solely depict a models performance as the differences are averaged out due to the high threshold cutoff. Moreover, it is used to prove the stability of the model but rarely used as the actual sampling rate. As evident, that only Arrihythmia has more than 10\% positive negative rate.}
\resizebox{\textwidth}{!}{
\begin{tabular}{|ccc ccc cccc|}
 \toprule
 Data set & \makecell{Isolation\\ Forest} & \makecell{One Class\\ SVM} & \makecell{Elliptic \\Envelope} & LOF & GOAD & \makecell{ICL} &  COPOD & RecTrans & \makecell{SORTAD\\ (ours)}\\ 
 \hline
 \midrule
 SMTP & \textbf{8.94} & 8.68 & 8.62 & 5.63 & 7.22 & 7.32 & 7.56 & 8.71 & 8.54  \\ \hline
 Mammography & 6.17 & 3.60 & 6.86 & 5.98 & 4.99 & 2.43 & \textbf{7.92} & 7.63 & 6.96  \\ \hline
 Credit Card & 6.89 & 8.00 & 4.74 & 1.01 & 5.75 & 2.83 & 8.36 & \textbf{12.63} & 8.05  \\ \hline
 Thyroid & 9.22 & 10.41 & 9.49 & 6.13 & 7.86 & 3.19 & 7.24 & 6.33 & \textbf{11.65}  \\ \hline
 Forest Cover & 6.06 & 8.95 & 1.20 & 8.23 & 7.46 & 8.64 & 4.96 & 4.58 & \textbf{9.75}  \\ \hline
Arrhythmia & 4.40 & 4.57 & 2.63 & 1.71 & 4.28 & 1.96 & 3.74 & \textbf{4.83} & 4.43  \\ \hline
Shuttle & 9.65 & 9.65 & 9.60 & 9.16 & 9.74 & 9.44 & 9.79 & 9.68 & \textbf{10.07}  \\ \hline
Pendigits & 7.58 & 9.26 & 3.46 & \textbf{9.43} & 3.26 & 3.01 & 5.91 & 9.03 & 8.31 \\ \hline
Annthyroid & 3.89 & \textbf{6.89} & 5.52 & 5.30 & 4.33 & 4.00 & 2.76 & 4.63 & 5.30  \\ \hline
Vowels & 4.16 & 4.29 & 1.68 & 7.91 & \textbf{8.06} & 1.17 & 1.86 & 3.57 & 3.43 \\ \hline \hline
 mean$\pm$std & 6.70$\pm$2.03 & 7.43$\pm$2.33 & 5.38$\pm$3.00 & 6.05$\pm$2.73 & 6.30$\pm$1.97 & 4.40$\pm$2.80 & 6.01$\pm$2.48 & 7.16$\pm$2.73 & \textbf{7.65$\pm$2.49} \\ 
 \bottomrule
\end{tabular}
}
\label{table_vs_random_10}
\end{table*}

\begin{table*}[ht]
\centering
\caption{ROC-AUC results on specific data sets. SORTAD has the second best average score, since ROC-AUC is only used to evaluate the stability of the model and doesn't indicate which is the best model, since lower percentage thresholds will be used and not the whole range as ROC-AUC. Especially in imbalance data sets were the results tend to be more saturated.}
\resizebox{\textwidth}{!}{
\begin{tabular}{|ccc ccc cccc|}
 \toprule
 Data set & \makecell{Isolation\\ Forest} & \makecell{One Class\\ SVM} & \makecell{Elliptic \\Envelope} & LOF & GOAD & \makecell{ICL} &  COPOD & RecTrans & \makecell{SORTAD\\ (ours)}\\ 
 \hline
 \midrule
 SMTP & 0.937 & 0.975 & 0.963 & 0.808 & 0.748 & 0.797 & 0.925 & 0.972 & \textbf{0.979}  \\ \hline
 Mammography & 0.887 & 0.818 & 0.885 & 0.872 & 0.543 & 0.717 & \textbf{0.924} & 0.896 & 0.874  \\ \hline
 Credit Card & 0.940 & 0.929 & 0.839 & 0.553 & 0.853 & 0.935 & 0.939 & 0.873 & \textbf{0.944}  \\ \hline
 Thyroid & 0.967 & 0.985 & 0.979 & 0.945 & 0.894 & 0.952 & 0.922 & 0.913 & \textbf{0.991}  \\ \hline
 Forest Cover & 0.908 & 0.951 & 0.692 & \textbf{0.990} & 0.921 & 0.969 & 0.881 & 0.798 & 0.985  \\ \hline
Arrhythmia & 0.809 & 0.843 & \textbf{0.845} & 0.690 & 0.809 & 0.734 & 0.818 & 0.762 & 0.794  \\ \hline
Shuttle & 0.997 & 0.991 & 0.990 & 0.997 & 0.992 & \textbf{1.00} & 0.995 & 0.991 & 0.993  \\ \hline
Pendigits & 0.950 & 0.965 & 0.852 & \textbf{0.996} & 0.889 & 0.979 & 0.920 & 0.959 & 0.955\\ \hline
Annthyroid & 0.802 & \textbf{0.956} & 0.922 & 0.904 & 0.489 & 0.909 & 0.784 & 0.787 & 0.872  \\ \hline
Vowels & 0.738 & 0.771 & 0.689 & 0.941 & 0.920 & \textbf{0.997} & 0.512 & 0.652 & 0.678
\\ \hline \hline
 mean$\pm$std & 0.894$\pm$0.080 & \textbf{0.918$\pm$0.074} & 0.866$\pm$0.102 & 0.870$\pm$0.140 & 0.806$\pm$0.158 & 0.899$\pm$0.158 & 0.862$\pm$0.130 & 0.860$\pm$0.103 & 0.907$\pm$0.099 \\ 
 \bottomrule
\end{tabular}
}
\label{table_recall_auc}
\end{table*}

\clearpage
\subsection{Disclaimer}
https://edc.intel.com/content/www/us/en/products/performance/benchmarks/overview/

\end{document}